\documentclass{article}
\usepackage{ijcai25}
\usepackage{amsmath}
\usepackage{times}
\usepackage{soul}
\usepackage{url}
\usepackage[hidelinks]{hyperref}
\usepackage[utf8]{inputenc}
\usepackage[small]{caption}
\usepackage{graphicx}
\usepackage{amsmath}
\usepackage{amsthm}
\usepackage{booktabs}
\usepackage{algorithm}
\usepackage{algorithmic}
\usepackage[switch]{lineno}
\usepackage{pgfplots}
\usepackage{amsfonts}
\usepackage{multirow}
\title{Empowering Vision Transformers with Multi-Scale Causal Intervention for Long-Tailed Image Classification}

\author{
Xiaoshuo Yan$^1$
\and
Zhaochuan Li$^2$\and
Lei Meng$^{1}$\and
Zhuang Qi$^1$\and
Wei Wu$^1$\and
Zixuan Li$^1$\and
Xiangxu Meng$^1$\\
\affiliations
$^1$School of Software, Shandong University, Jinan, China\\
$^2$Inspur, Jinan, China\\
\emails
\{yanxiaoshuo, z\_qi, wu\_wei, lizixuan0707\}@mail.sdu.edu.cn,
lizhaoch@inspur.com,
\{lmeng, mxx\}@sdu.edu.cn
}

\begin{document}

\maketitle

\begin{abstract}
Causal inference has emerged as a promising approach to mitigate long-tail classification by handling the biases introduced by class imbalance. However, along with the change of advanced backbone models from Convolutional Neural Networks (CNNs) to Visual Transformers (ViT), existing causal models may not achieve an expected performance gain. This paper investigates the influence of existing causal models on CNNs and ViT variants, highlighting that ViT's global feature representation makes it hard for causal methods to model associations between fine-grained features and predictions, which leads to difficulties in classifying tail classes with similar visual appearance. To address these issues, this paper proposes TSCNet, a two-stage causal modeling method to discover fine-grained causal associations through multi-scale causal interventions. Specifically, in the hierarchical causal representation learning stage (HCRL), it decouples the background and objects, applying backdoor interventions at both the patch and feature level to prevent model from using class-irrelevant areas to infer labels which enhances fine-grained causal representation. In the counterfactual logits bias calibration stage (CLBC), it refines the optimization of model's decision boundary by adaptive constructing counterfactual balanced data distribution to remove the spurious associations in the logits caused by data distribution. Extensive experiments conducted on various long-tail benchmarks demonstrate that the proposed TSCNet can eliminate multiple biases introduced by data imbalance, which outperforms existing methods.

\end{abstract}

\section{Introduction}
Real-world data typically follows long-tailed distributions, resulting in models that primarily optimize for head classes and demonstrate limited generalization to tail classes \cite{zhang2023deep}. Existing CNN-based long-tailed algorithms including class balancing methods \cite{cui2019class,ren2020balanced}, data augmentation \cite{wang2024kill,ahn2022cuda}, enhanced training strategies \cite{wang2021contrastive,du2023global}. With the development of Transformers, ViT employs an attention-based global feature extraction approach, facilitating the capture of finer-grained features relative to CNN architectures. However, this does not change the essence of the model's reliance on statistical information from the data, leading to an overall performance gain but a persistent gap between head and tail class performance, leaving the long-tail problem unresolved.
\begin{figure}[t]
\centering
\includegraphics[width=0.47\textwidth]{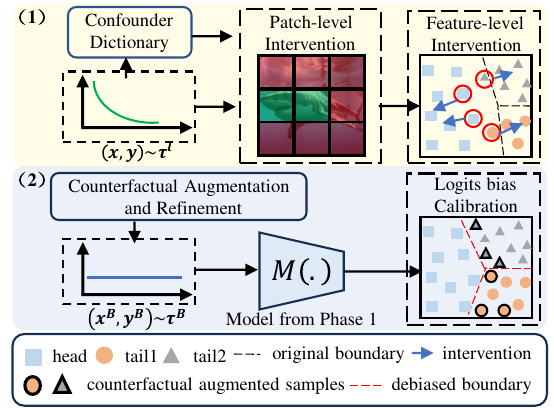}
\caption{The illustration of the proposed TSCNet. It removes semantic bias through hierarchical causal intervention to enhance the causal representation of tail classes. In the second stage, it adaptively calibrates logit bias through counterfactual intervention.}
\vspace{-0.25cm}
\label{fig:motivation}
\end{figure} 
Long-tail image classification in ViT can be improved using two main approaches: parameter-efficient fine-tuning strategies and information enhancement. The former method \cite{li2024improving} mainly leverages pre-trained knowledge to enhance the generalization of tail classes. LPT \cite{dong2022lpt} designs visual prompts for group-wise categories to improve the learning of unique representations for tail classes, while LIFT \cite{shi2023parameter} adopts a partial parameter fine-tuning approach to enhance the discriminative ability for tail classes. However, they struggle with tail classes that have high intra-class complexity by fine-tuning with a limited number of parameters. Information enhancement methods aim to augment the information for tail classes. VL-LTR \cite{tian2022vl} leverages textual features to enhance the learning of image features, and DeiT \cite{rangwani2024deit} extracts information from pre-trained CNNs through knowledge distillation. However, these methods often struggle to obtain accurate knowledge of tail classes, leading to semantic confusion between tail and head classes. The key to solving these problems lies in learning highly relevant visual features and mitigating spurious correlations caused by long-tailed distribution, which can be achieved through causal inference methods. However, directly applying existing causal methods \cite{tang2020long,zhu2022cross} to ViT fails to yield performance gains akin to CNN-based models due to they are difficult to model the spurious association between the fine-grained features extracted by ViT and the predictions by calibrating the logits with the estimated category consistency bias. This leads to the problem that existing causal methods struggle to eliminate the spurious associations between a woman's image and related categories like "girl" or "table," with most of the misclassified tail categories being confused with similar categories, as shown in Figure \ref{fig:figure2}.

To address these issues, this paper proposes a two-stage causal modeling framework by multi-scale causal intervention termed TSCNet, as shown in Figure \ref{fig:motivation}. To enhance the model's fine-grained causal representation and mitigate the spurious associations on logits, we design two stages: hierarchical causal representation learning (HCRL) and counterfactual logits bias calibration (CLBC). HCRL enhances the model's fine-grained causal representation for tail classes by introducing class-independent semantic information such as background at both the patch-level and global feature-level. This enables the model to focus on class-relevant regions through hierarchical interventions. CLBC calibrates the spurious associations in label predictions caused by domain distribution from counterfactual perspective. By counterfactual generation and adaptively refining the intensity of counterfactual augmentation to construct different distributions, we effectively model category relationships and calibrate logits' bias caused by long-tailed distribution. The two-stage causal modeling method benefits from the sparse mechanism shift (SMS), enabling independent interventions on multiple biases. TSCNet preserves the performance of head classes while striving for higher performance in tail class.

% To address these issues,  first established a structural causal model (SCM) and, which solves the semantic confusion in feature representation and the data distribution bias in logits,  The first stage is the , where we enhance the model's fine-grained causal representation learning for tail classes by introducing the patch-level and global feature-level intervention to distinguish causal semantics. The second stage is the counterfactual logits bias calibration (CLBC). Starting from the higher-level counterfactual perspective, we propose a model-agnostic counterfactual method to calibrate logits bias. By counterfactual generation and adaptively refining the intensity of counterfactual augmentation to construct different distribution, we effectively model category relationships and calibrate logits bias. The two-stage causal modeling method benefits from the sparse mechanism shift(SMS) in causal inference, enabling independent interventions on multiple biases. This approach potentially preserves the performance of head classes while striving for higher performance in fine-grained tail relationships.

Experiments were conducted on two datasets, including performance comparison, ablation study, case study, and other in-depth analyses. The results confirm that TSCNet effectively enhances causal representations for tail data, while mitigating the logits bias caused by long-tailed distributions. The main contributions of this paper are:
\begin{itemize}
\item{This paper points out that due to the different feature extraction, existing long-tailed causal methods face challenges when applied to transformer architectures.}
%我们提出了一个两阶段因果建模框架，从因果关系的角度对长尾分布进行建模。据我们所知，这是可以有效适用于Transformer的因果解决方案。
%
\item{This paper propose a two-stage causal modeling framework by multi-scale causal intervention. To the best of our knowledge, it is the first causal framework that uses backdoor adjustment to remove various biases and is applicable to ViT.} 

\item{Experimental results demonstrate that TSCNet highlight the ability to address multiple biases, simultaneously reducing erroneous predictions for both head and tail classes caused by semantic bias and distributional bias.

}
\end{itemize}

\begin{figure}[t]
\centering
\includegraphics[width=0.47\textwidth]{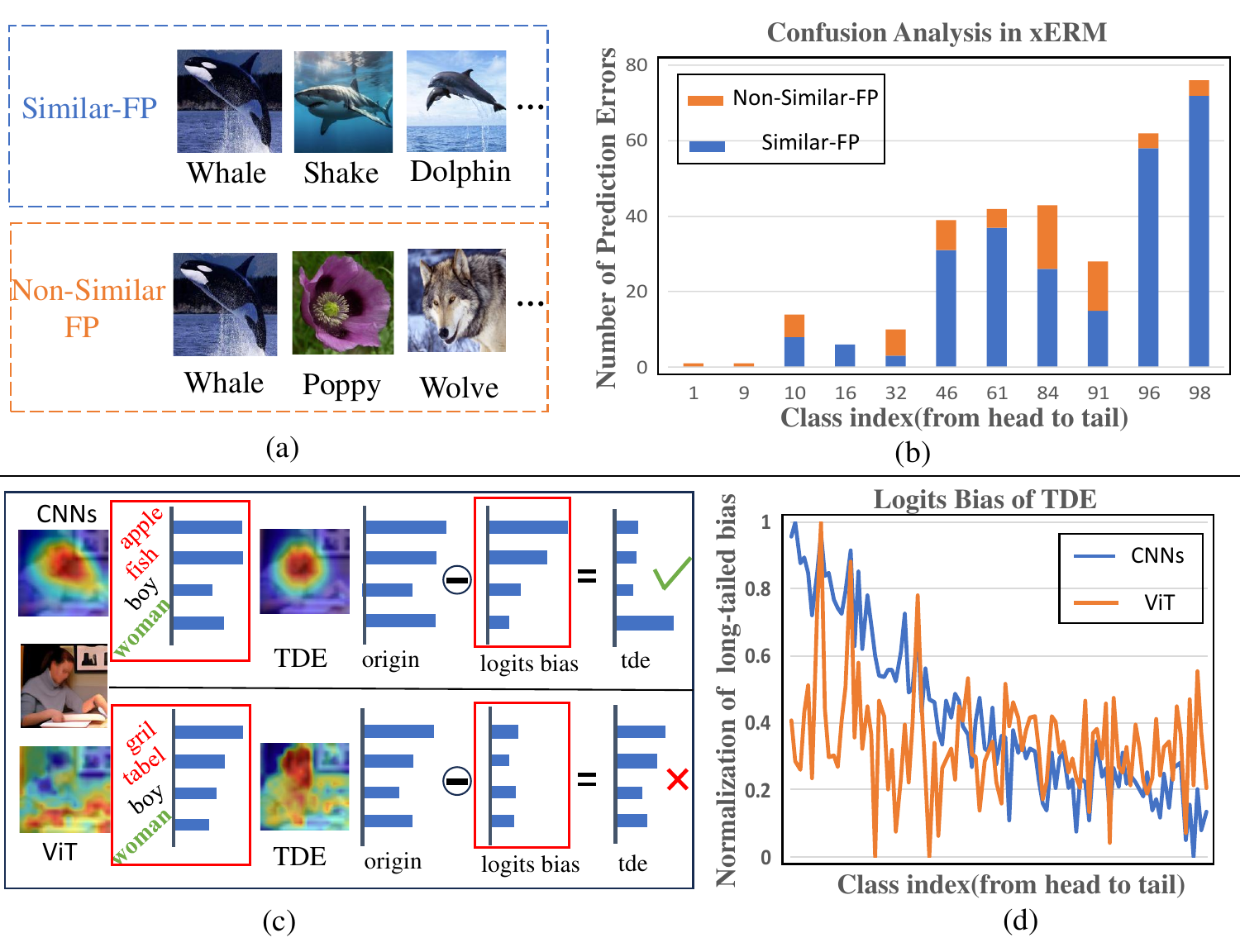}

                          \caption{The example of Similar-FP and Non-similar-FP on (a), the error confusion analysis of the existing causal method xERM on ViT on (b), the bias correction mechanism of causal methods TDE on (c), the difference in counterfactual estimation logits class consistency long-tail bias between the method TDE on ViT and CNNs. on (d).}
\vspace{-0.25cm}
\label{fig:figure2}
\end{figure} 

\section{Related Work}
\subsection{Long-tailed Image Classification}

%长尾方案分为两类 一类描述CNN上的长尾方法发展：四类方法：重平衡,训练策略优化,信息增强,因果推断, 一类介绍在ViT上的长尾方法分类：主要集中在对于ViT的微调策略：高斯优化，提示微调，参数高效微调

%早期的研究主要致力于在训练过程中实现类别间的平衡优化，常通过设计采样策略，或者设计加权损失函数来增强模型对于尾部类别的优化，这些方法造成了头部类性能的严重下降。随着研究的深入，长尾图像分类的方法逐渐结合了训练策略改进和信息增强技术，以缓解类别不平衡问题。这些方法包括解耦表示学习和分类器来校准分类器的偏差，元学习策略优化尾部类学习，设计具有不同分支的集成学习方法，以及通过课程学习增强尾部类数据和迁移学习方法迁移头部知识，从而增强模型的泛化能力。此外，因果推断方法也在长尾分类中发挥了重要作用。研究表明，基于因果推理的长尾方法在卷积神经网络（CNN）上展现了优越的性能，对于去除长尾学习偏差展现出有效性。

%针对于长尾分布问题，先前的研究主要集中在三个不同的方面:类平衡方法，它通过设计重采样策略,重加权损失函数,logits调整等来增强对于尾部类的优化。但这类方法通常导致头部类性能的大幅下降。第二类方法为数据增强方法，通过课程学习选取图像增强策略，将头部类知识到尾部类的迁移学习，以及尾部类的特征增强等方法来改善尾部类信息不足的问题。第三类方法为训练策略改进，该类方法多通过解耦表示学习和分类器，集成学习策略等来进一步改善对于头尾部的优化问题。
Previous studies addressing the negative impacts of long-tail distributions have focused on three distinct aspects: class balancing methods, which enhance optimization for tail classes by designing resampling strategies \cite{mahajan2018exploring,cui2019class}, reweighting loss functions \cite{cao2019learning,ren2020balanced,zhoupareto}, and adjusting logits \cite{hong2021disentangling}; data augmentation, which improve the information scarcity of tail classes through curriculum learning for image augmentation\cite{ahn2022cuda,wang2024kill}, transfer learning from head classes to tail classes \cite{chen2023transfer}, and feature enhancement for tail classes \cite{li2024feature}; improving training strategies \cite{wang2021contrastive,du2023global}, which typically involve decoupling representation learning from classifiers \cite{kang2019decoupling} and employing ensemble learning strategies \cite{zhou2020bbn,cui2022reslt} to further optimize both head and tail classes. Causal methods \cite{tang2020long,zhu2022cross} have shown remarkable performance improvements in CNNs by calibrating the logits bias induced by data distribution through backdoor adjustments. Transformer-based long-tailed methodologies\cite{xu2023learning,zhu2024generalized} primarily focus on fine-tuning strategies for ViT, including prompt tuning to enhance shared prompts for tail classes\cite{dong2022lpt,li2024improving},  parameter-efficient fine-tuning techniques\cite{shi2023parameter} to facilitate the learning of tail classes. Some methods further enhance the representation of tail classes by incorporating external knowledge through visual-language contrastive learning\cite{tian2022vl} and knowledge distillation techniques\cite{rangwani2024deit}.
\subsection{Causal Inference in Image Classification}
%基于图像分类的因果方法分为三种(针对于ViT)：前门，后门，反事实
%现有的因果理论在图像分类领域已经比较成熟，已有研究通过后门调整策略，通过设计因果分类器或注意力机制以及反事实推理来识别和减轻上下文、背景\cite{yang2023context}等混杂因素的干扰，可以有效提升图像分类及相关任务的性能；此外，在没有明确的混淆因素情况下，现有的前门调整策略则设计局部-局部特征和局部-全局特征注意力来获得图像分类可辩别的因果特征。此外还有因果不变表示学习的方法通过风格生成模型或者傅里叶变换技术，同时设计不变损失来增强对于图像因果因素的识别。

Causal inference and counterfactual reasoning have received increasing attention in a variety of tasks in computer vision, including scene graph generation\cite{sun2023unbiased}, image recognition\cite{wang2021causal}, and video analysis\cite{wang2024modeling}. Causal methods in the field of image classification demonstrated significant performance improvements. Existing research has implemented backdoor adjustment strategies by designing causal classifiers\cite{liu2022contextual}, using attention mechanisms\cite{yang2023context} to identify and mitigate the interference of confounding factors\cite{zhang2024bi}. Moreover, prevailing front-door adjustment strategies involve designing local-global feature attention mechanisms\cite{yang2021causal,liu2023cross} to extract distinguishable causal features. Additionally, causal invariant representation learning methods \cite{mao2022causal,liu2024causality} utilize style generation models or Fourier transform techniques in conjunction with invariant loss functions\cite{lv2022causality} to improve the identification of causal factors. However, existing causal methods in image classification are inadequate for addressing the long-tailed distribution problem, as they overlook the influence of long-tailed bias on causal graph construction and the limited effectiveness of interventions in tail classes with sparse data.

\section{Problem Formulation}
The long-tailed dataset is represented as ${\mathcal{D}}=\{x,y\}$. Let $n_j$ denote the number of training sample for class $j$, and let $n = \sum_{j=1}^{C} n_j$ be the total number of training sample and $n_1 \gg n_C$. Conventional methods extract visual features : $F_v= M_v(x)$, where $M_v(\cdot)$ denotes the feature extractor. Then, predicting the category of the sample, i.e. $P = classifier(F_v)$. Different from conventional methods, in the first stage of HCRL, we initially extract confounder $S=[s_1,s_2, ... s_n]$ using object detection models. Subsequently, we perform causal interventions $P(Y|do(X))$ at both the token level and the global feature level, thereby obtaining a causally enhanced model $M(\cdot)$. In the second stage, we construct a counterfactual balanced distribution $\overline{x}$ by augmenting deficient tail class samples through counterfactual data augmentation $F(\overline{x},L^e_c)$ while adaptively adjusting the intensity $L^e_c$ of the augmentation to perform causal interventions $P(Y|do(D))$. This process can get predictions after calibration $P_c = M(\overline{x})$.

%得到校准后的预测
\section{Method}
This study proposes a two-stage debiasing method for long-tail learning, called TSCNet. The method constructs a structured causal graph to analyze the interfering factors in the inference path. Specifically, TSCNet consists of two main stages, as shown in Figure \ref{fig:framework}. The Hierarchical Causal Representation Learning (HCRL) stage enhances the model's fine-grained causal representation through debiasing at the patch and global feature levels eliminating semantic confusion. The Counterfactual Bias Calibration (CLBC) stage utilizes counterfactual data augmentation and refinemention strategies to reduce the logits' bias caused by data distribution.
\subsection{Causal View at Long-tailed Classification}
\begin{figure}[t]
\centering
\includegraphics[width=0.47\textwidth]{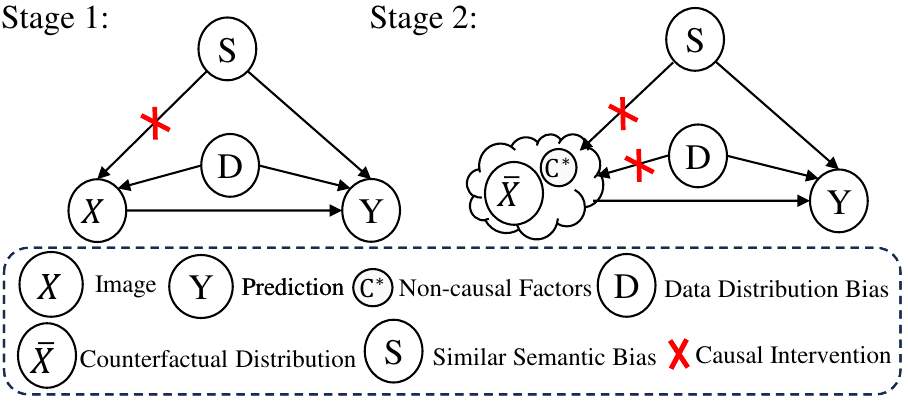}
\caption{The Causal view of Long-tailed Image Classification.}
\vspace{-0.3cm}
\label{fig:causal}
\end{figure} 
We use the structural causal model to model the variable relationships of complex spatiotemporal data in long-tailed image classification tasks. As illustrated in Figure \ref{fig:causal}, it is a directed acyclic graph $\mathcal{G} = \{N, E\} $ in which the nodes N denote variables and edges E denote association between variables.
The SCM $\mathcal{G}$ includes four variables:  image $X$, semantic confounder $S$, data distribution confounder $D$ and prediction $Y$. The correlations in graph $\mathcal{G}$ are as follows:

\begin{figure*}[t]
\centering
\includegraphics[width=0.97\textwidth]{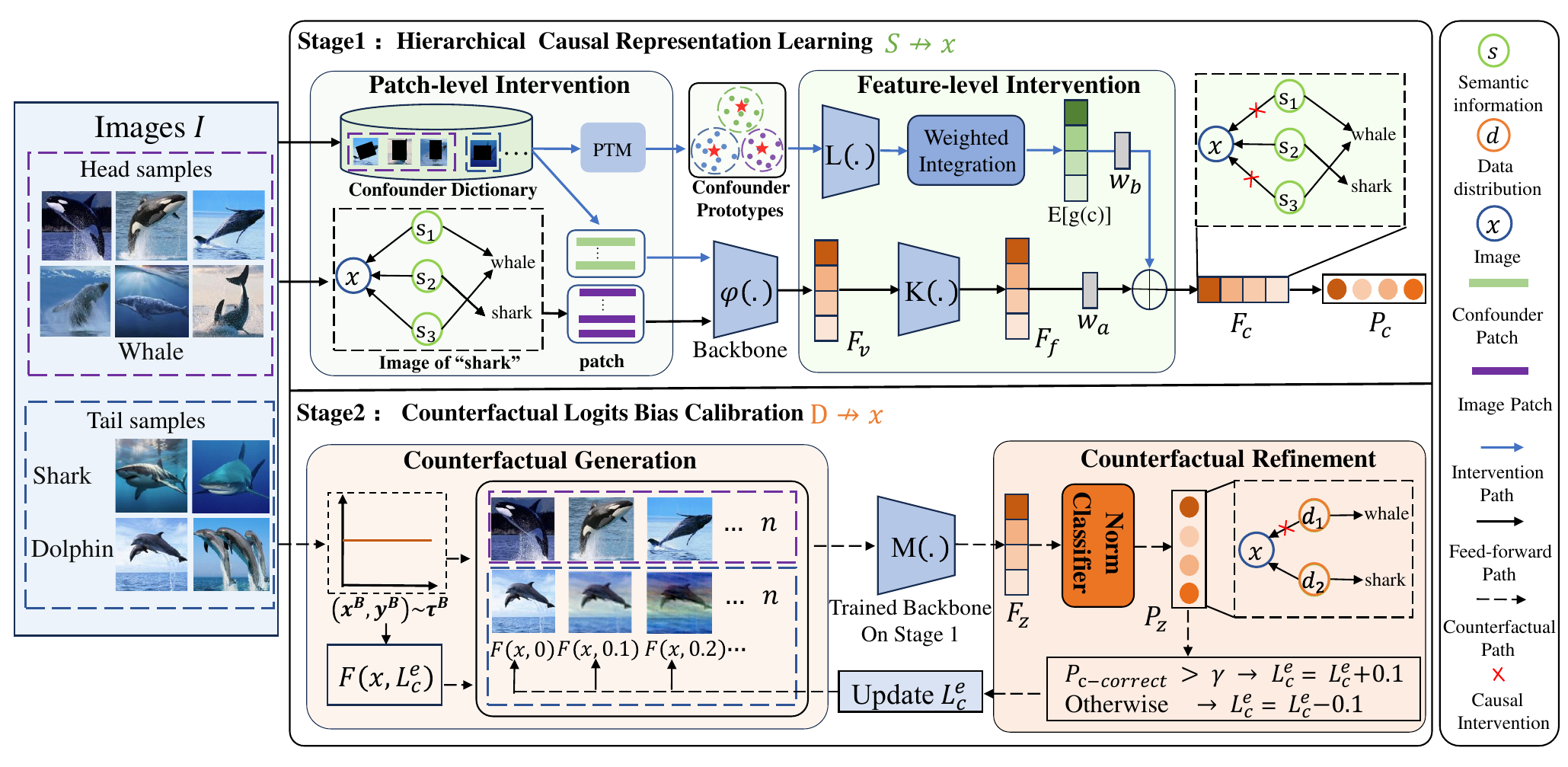}
\caption{Illustration of the proposed TSCNet. It contains two main stages: HCRL and CLBC. The former introduces class-independent semantic information and performs backdoor adjustments to enhance the model’s fine-grained causal representation for tail classes $S\not\rightarrow X$. The latter generates counterfactual distribution to calibrate logits bias and model category relationships $D\not\rightarrow X$. }
\vspace{-0.2cm}
\label{fig:framework}
\end{figure*}

% $\boldsymbol{x_t \xrightarrow{} v_t}$, $\boldsymbol{x_h \xrightarrow{} v_h}$. This path indicates that the model extracts features based on the image content.

$\boldsymbol{S \xrightarrow{} X}$. This path indicates that similar semantic factors such as the background influence the composition of image content and tend to affect the tail classes where data is sparse.

% $\boldsymbol{B \xrightarrow{} v_t}$, $\boldsymbol{B \xrightarrow{} v_h}$. This path indicates that Background factors influence the feature distribution, exacerbating the disparity in feature extraction for tail class data.
$\boldsymbol{S \xrightarrow{} Y}$. This path indicates that the predicted label distributions(logits) follow their own training domain prior.

$\boldsymbol{D \xrightarrow{} X}$. This path indicates that an image is sampled according to the selected data distribution,e.g., imbalanced data distribution is prone to head classes.

$\boldsymbol{D \xrightarrow{} Y}$. This path indicates that the predicted label distributions follow their own training domain prior.

$\boldsymbol{X \leftarrow{} S \xrightarrow{} Y}$, $\boldsymbol{X \leftarrow{} D \xrightarrow{} Y}$. This two back-door path contribute spurious correlation between $X$ and $Y$, where $S$ and $B$ acts as confounder. 
\subsection{Hierarchical Causal Representation Learning}
To eliminate the semantic confusion between tail classes and head classes, we propose a hierarchical causal representation learning method. By extracting class-agnostic information and performing causal interventions at both the patch and global feature levels, the method enhances the model's fine-grained representation learning for tail classes. 

Mitigating the bias caused by $S$ is to intervene on $X$, ensuring that class-agnostic semantic information contributes equally to the image classification. We extract class-agnostic semantic information and intervene on $X$:
\begin{equation}
\setlength\abovedisplayskip{0.2cm}
\setlength\belowdisplayskip{0.2cm}
\resizebox{0.7\hsize}{!}{$
P(Y \mid do(X))=\sum_S P(Y \mid X, S) P\left(S\right)$}
\end{equation}
where $do(X)$ denotes intervene on $X$, the path in Fig. 3 from $S$ to $X$ is cut-off. Due to the inability to combine all class-agnostic information with the image, only approximate interventions are possible. We propose a hierarchical intervention strategy to strengthen the intervention for tail class and improve the fine-grained causal representation of tail classes.

\subsubsection{Patch-level Intervention}
At the patch-level intervention, we introduce class-agnostic patch information alongside the original image patches, leveraging the encoder's attention mechanism to finely uncover the causal regions within the image. Specifically, we can apply this method on CNNs to merge the original image with theclass-agnostic information.

\textbf{Confounder Dictionary S} 
 We extract class-irrelevant semantic information from the training set and construct a confounder dictionary. Given an image $x_i$, we use off-the-shelf methods such as object detection model or Grad-CAM \cite{selvaraju2020grad} to detect the main subject of the image.  Then, we apply an inverse transformation to obtain the class-agnostic information mask $M_i$. Then we paste the mask onto the original image by $M_i \odot x_i$ to obtain a class-irrelevant image $s_i$.Then, we construct the confounder dictionary $S=[s_1,s_2,...,s_n]$.

In theory, patch-level interventions require multiple backpropagations through all class-irrelevant information. To reduce training overhead, we assume that the distribution of confounding patches is uniform and adopt a random sampling strategy to intervene:
\begin{equation}
\setlength\abovedisplayskip{0.2cm}
\setlength\belowdisplayskip{0.2cm}
\resizebox{1\hsize}{!}{$
P(Y\mid\mathrm{do}(X))\approx\frac {1}{N}\sum_{j=1}^N P(Y\mid f(X,s_j))\approx P(Y\mid f(E(X),E(s_k)))$}
\end{equation}
where $s_k$ is randomly sampled from the confounder dictionary $S$, $E(\cdot)$ is the patch embeddings of the image, and $f(x, s_k)$ is the function that indicates stacking and concatenation on the sequence length dimension, the length of the sequence can be changed by random sampling from $E(s_k)$.

Then,we can get the causal-enhanced representation:
\begin{equation}
    F_v = \varphi(f(E(X),E(s_k)))
\end{equation}
where $\varphi(.)$ is the visual backbone.

\subsubsection{Feature-level Intervention}
To further mitigate the effects caused by class-agnostic semantics on the feature distribution and enhance the model’s generalization ability for tail classes, we introduce a global feature causal intervention module.

We construct a confounder prototype dictionary 
$S_p=[c1,c2,...,c_l]$ structured as an $l*d$ matrix to systematically address these feature-level factors. Where $l$ is the dictionary size and $d$ is the feature dimension using the pre-trained backbone. We apply k-means++ \cite{bahmani2012scalable} to derive $c_i$ from the confounder dictionary $S$, each $c_i$ is the average feature of its cluster.
To implement the theoretical interventions in Eq 2 and reduce computation, we use Normalised Weighted Geometric Mean (NWGM) \cite{xu2015show} to approximate the results expected from the above feature layers:
\begin{equation}
\setlength\abovedisplayskip{0.2cm}
\setlength\belowdisplayskip{0.2cm}
\resizebox{0.7\hsize}{!}{$
    P(Y\mid\mathrm{do}(X))\overset{\mathrm{NWGM}}{\operatorname*{\approx}}\sum_S P(Y \mid X, S) P\left(S\right)$}
\end{equation}
We parameterize the network model to approximate the conditional probability of Eq.6, inspired by[48], as follows:
\begin{equation}
\setlength\abovedisplayskip{0.2cm}
\setlength\belowdisplayskip{0.2cm}
\resizebox{0.65\hsize}{!}{$
    P(Y\mid do(X))=W_aF_f+W_b\mathbb{E}_c[g(c)]$}
\end{equation}
where $W_{a}\in\mathbb{R}^{d_{m}\times d_{a}}$ and $W_{b}\in\mathbb{R}^{d_{m}\times d}$ are learnable parameters. We approximate $\mathbb{E}_c[g(c)]$ as a weighted integration of all background prototypes:
\begin{equation}
\setlength\abovedisplayskip{0.2cm}
\setlength\belowdisplayskip{0.2cm}
\resizebox{0.5\hsize}{!}{$
\mathbb{E}_c[g(c)]=1/N_i\sum_{i=1}^N\mu_ic_i$}
\end{equation}
where $\mu_i$ represents the important weight coefficient measuring the interaction between each $c_i$ and the feature $F_v$.

\subsection{Counterfactual Logits Bias Calibration }
Although we obtained features of causal enhancement in the first stage, the model relies on this long-tail distribution prior, leading to bias in label prediction. Therefore, we propose a model-agnostic counterfactual intervention method, which generates a balanced distribution through counterfactual augmentation and refinement to adaptively calibrate the logits:
\begin{equation}
\setlength\abovedisplayskip{0.2cm}
\setlength\belowdisplayskip{0.2cm}
\resizebox{0.65\hsize}{!}{$
P(Y|do(X))=\sum_{D}P(Y|X,D)P(D)$}
\end{equation}
where $d=0$ denotes the imbalanced 
data distribution, and $d=1$ denotes the balanced data distribution. 

\subsubsection{Counterfactual Generation}
Simple balanced sampling is not applicable as it leads to the model's overfitting to tail classes. We perform counterfactual data augmentation by disturbing non-causal factors to construct a counterfactual balanced distribution.

We use Fourier transformation \cite{lv2022causality} for counterfactual data augmentation, which has the desirable property of preserving the high-level semantics of the original signal, while the amplitude component contains low-level statistics, and it also helps save computational cost:
\begin{equation}
\setlength\abovedisplayskip{0.2cm}
\setlength\belowdisplayskip{0.2cm}
\resizebox{0.4\hsize}{!}{$
    \mathcal{F}(x)=\mathcal{A}(x)\times e^{-j\times\mathcal{P}(x)}$}
\end{equation}
where $\mathcal{A}(x)$, $\mathcal{P}(x)$ denote the amplitude and phase components respectively. We then perturb the amplitude information via linearly interpolating between the amplitude spectrums of the original image $x$ and an image $x'$ sampled randomly:
\begin{equation}
\setlength\abovedisplayskip{0.2cm}
\setlength\belowdisplayskip{0.2cm}
\resizebox{0.6\hsize}{!}{$
    \hat{\mathcal{A}}(x^o)=(1-\lambda)\mathcal{A}(x^o)+\lambda\mathcal{A}((x^{\prime})^o)$}
\end{equation}
where $\lambda\sim U(0, L_{c}^{e})$ and $L_{c}^{e}$ controls the strength of perturbation, which adjusts the perturbation strength for class $c$ during an epoch $e$. Then we can obtain the counterfactual augmented image. We can obtain the counterfactual augmented image:
\begin{equation}
\setlength\abovedisplayskip{0.2cm}
\setlength\belowdisplayskip{0.2cm}
\resizebox{0.75\hsize}{!}{$
\mathcal{F}(x^{a})=\hat{\mathcal{A}}(x^{o})\times e^{-j\times\mathcal{P}(x^{o})},x^{a}=\mathcal{F}^{-1}(\mathcal{F}(x^{a}))$}
\end{equation}
\subsubsection{Counterfactual Refinement}
We adaptively adjust the strength $L_c^e$ of counterfactual data augmentation to enable the model to progressively adjust the logits bias from easy to difficult. At epoch 
$e$, we define a computation function $P_{l}$ for each class to adaptively update the perturbation strength:
\begin{equation}
\setlength\abovedisplayskip{0.2cm}
\setlength\belowdisplayskip{0.2cm}
\resizebox{0.5\hsize}{!}{$
L_c^e=P_{l}(\mathcal{D}_c,L_c^{e-1},M(\cdot),\gamma)$}
\end{equation}
where $\gamma$ is threshold hyperparameter, $M(\cdot)$ is the model from stage 1. We can update $V_{\mathrm{LoL}}$ as follows: 
\begin{equation}
\setlength\abovedisplayskip{0.2cm}
\setlength\belowdisplayskip{0.2cm}
\resizebox{0.65\hsize}{!}{$
    P_{l} = L_{c}^{e-1}+0.1\quad\mathrm{if~}Acc(\mathcal{D}_{c},M(\cdot))\geq \gamma$}
\end{equation}
\begin{equation}
\setlength\abovedisplayskip{0.2cm}
\setlength\belowdisplayskip{0.2cm}
\resizebox{0.46\hsize}{!}{$
    V_{\mathrm{LoL}} = L_{c}^{e-1}-0.1\quad{\mathrm{otherwise}}$}
\end{equation}
where $Acc$ is a function which outputs the number of correctly predicted examples by the model $f_\theta$.

After updating $L_c^e$, TSCNet control the intensity of generating the non-causal factor $\hat{\mathcal{A}}(x^{o})$ and conterfactual augmented image $x^a$ to construct a balanced dataset $\overline{x} =(x^B,y^B){\sim}\tau^B$. Then, We can get $P_z = M(\overline{x})$.

\subsection{Training Strategies}
The training of TSCNet follows two steps: the first step is de-confounded training for HCRL. After obtaining a causal representation-enhanced model, counterfactual fine-tuning is subsequently applied for CLBC. The details are as follows:

The HCRL stage uses causal intervention modules at the patch and feature levels. The process is constrained by:
     \begin{equation}
     \setlength\abovedisplayskip{0.2cm}
\setlength\belowdisplayskip{0.2cm}
\resizebox{0.46\hsize}{!}{$
     \mathcal{L}_{cls}=-(\sum\nolimits_{i=1}^{C} y_i \log(\hat{y}_i))$}
    \end{equation}
    % \item{\textbf{Step 2 (refining tag predicitons with pre-learned hierarchy and constructing  heterogeneous graph):}} Freezing Cascaded Spatial Net, matching visual regions $\textbf{r}$  and visual prototypes $\textbf{p}$, and then  refining the tagging predictions $\hat{\textbf{t}}$.
 
The CLBC construct and refine counterfactual distribution $\overline{x}$ to mitigate the long-tail bias. The process is constrained by:
     \begin{equation}
     \setlength\abovedisplayskip{0.2cm}
\setlength\belowdisplayskip{0.2cm}
\resizebox{0.7\hsize}{!}{$
    \mathcal{L}_{f} =\mathcal{L}_{cls} + \alpha_{gf}\frac{1}{N}\sum_{i=1}^N\|M(x)-M(x')\|_2^2$}
    \end{equation}
    where $\alpha_{gf}$ is the weight factor, $M(.)$ is the model from step 1, $x'$ is a counterfactually augmented sample of $x$.

\section{Experiments}

\subsection{Experiment Settings}
\begin{table*}[t]
\begin{center}

\small
\label{tab:performance}

\setlength{\tabcolsep}{1mm}{
\begin{tabular}{c|c|c|c|c|c|c|c|c|c|c|c|c|c}
\hline 
\multirow{2}*{\bf{Algorithms}}&
\multirow{2}*{\bf{Backbone}}&
\multicolumn{3}{c|}{\bf{CIFAR100-ratio0.01}}&
\multicolumn{3}{c|}{\bf{CIFAR100-ratio0.02}} &
\multicolumn{3}{c|}{\bf{CIFAR100-ratio0.1}}&
\multicolumn{3}{c}{\bf{VireoFood172-ratio0.02}} \\
\cline{3-14}

{} & {}&{\bf Acc@all} & {\bf Acc@h} & {\bf Acc@t} & {\bf Acc@all} & {\bf Acc@h} & {\bf Acc@t} & {\bf Acc@all} & {\bf Acc@h} & {\bf Acc@t}& {\bf Acc@all} & {\bf Acc@h} & {\bf Acc@t} \\
\hline

 ResNet50&{ResNet50}& 0.404  & 0.661  & 0.128  & 0.454  & 0.690 & 0.219  & 0.557  & 0.662 & 0.559 & 0.748  & 0.850 & 0.530
\\ 

  ViT &{ViT} & 0.795  & 0.929  & 0.613 & 0.817  & 0.932 & 0.712  & 0.885 & 0.926  & 0.836& 0.811 & 0.884 & 0.642   \\

 VPT&{VPT} & 0.812  & 0.930 & 0.657  & 0.840  & 0.941  & 0.748 & 0.895  & 0.929  & 0.862 & 0.826 & 0.895 & 0.658\\  \hline

% Feature Level Alignment
% cross-modal constraints: arch-d, cmrr
% feature-level alignment: atnet, ig-cman, msmvfa
% CCD &{ViT} & 0.809 & 0.925 & 0.655 & 0.825 & 0.930 & 0.843 & 0.887 & 0.924 & 0.840\\ 

% LGCAM & {ViT} & 0.816 &  0.932 & 0.661  & 0.827  & 0.933  & 0.726 & 0.875  & 0.924  & 0.824 & 0.817 & 0.882 & 0.634\\ 

CCIM & {ViT} & 0.779	& 0.934  & 0.567  & 0.829  & \textbf{0.945}  & 0.711 &0.888	& \textbf{0.931}	&	0.841& 0.826 & \textbf{0.896} & 0.657\\ 

CaDeT&{ViT} & 0.788  & 0.933    & 0.597    & 0.823 &	0.936 & 0.703 & 0.887 & 0.928 &0.842 & 0.812 & 0.887 & 0.649\\

 GOAT &{ViT} & 0.819 &  0.930 & 0.671  & 0.837  &0.941   &  0.736& 0.890 & 0.926 & 0.845 & 0.829 & 0.890 & 0.672\\ 
\hline
 TDE &{ResNet50} & 0.450  & 0.644 &  0.202 & 0.490  & 0.635  & 0.328 & 0.561  & 0.665 & 0.432 & 0.751 & 0.814 & 0.567  \\

 % CMO &{ResNet50} & 0.418  & 0.672 & 0.186  & 0.469 & 0.716 & 0.236 & 0.569  & 0.678 & 0.419 &  &  & \\

 xERM &{ResNet50} & 0.455  & 0.680 &0.174   & 0.492  & 0.670  & 0.296 & 0.575  & 0.691  & 0.426 & 0.770 & 0.835 & 0.572 \\

CUDA &{ResNet50} & 0.431  & 0.639 & 0.197  & 0.472  & 0.612  & 0.255  & 0.565  & 0.671  & 0.431& 0.764 & 0.823 & 0.550 \\

 PLOT &{ResNet50} & 0.445  & 0.640 &  0.219 & 0.487  & 0.607  & 0.330  & 0.573  & 0.680  & 0.443&0.769 &0.830& 0.573 \\ 
TDE &{ViT} & 0.803 & 0.937 & 0.624  & 0.836  & 0.930  & 0.737 & 0.887  & 0.924 & 0.840 & 0.810  & 0.880 & 0.651  \\
 % CMO &{ResNet50} & 0.418  & 0.672 & 0.186  & 0.469 & 0.716 & 0.236 & 0.569  & 0.678 & 0.419 &  &  & \\

 xERM &{ViT} & 0.799  & 0.930 & 0.615 & 0.834 & 0.946  & 0.720  & 0.888  & 0.926  & 0.841 &0.813  & 0.883 & 0.650 \\
 LiVT &{ViT} & 0.807 & 0.921  & 0.674  & 0.823  & 0.923  & 0.732 &0.885  & 0.924   &0.847 & 0.834 & 0.873 & 0.752\\ 
 Gpaco &{ViT} & 0.832  & 0.913  & 0.717  & 0.858  & 0.934  & 0.789 & 0.907 & 0.915   & 0.899 &0.831 & 0.875 & 0.746 \\
 
 H2T &{ViT} & 0.840  &  0.915 &  0.740 &0.832   & 0.919 & 0.731  & 0.887 &  0.916 & 0.853 & 0.798  & 0.630 & 0.876 \\ 
 LPT &{VPT} & 0.861 & 0.933  & 0.778  & 0.884  & 0.931 & 0.853  & 0.908 & 0.916  & 0.899 & 0.830 & 0.888 & 0.690\\ 
TSCNet &{ResNet50} & 0.472  & 0.640  & 0.258 & 0.510 & 0.674 & 0.331 & 0.590 & 0.657  & 0.487 & 0.790 & 0.832  & 0.598 \\ 
 TSCNet &{ViT} & 0.860  & 0.932  & 0.778 & 0.877  & 0.937  & 0.819    & 0.905  & 0.927  & 0.885 & 0.847 & 0.885 & 0.750 \\ 

 TSCNet &{VPT} & \textbf{0.887}  & \textbf{0.934}  & \textbf{0.830}  & \textbf{0.901} & {0.937}  & \textbf{0.877}  & \textbf{0.915} & {0.924}  & \textbf{0.917} &\textbf{0.875} & 0.890 & \textbf{0.819}\\
\hline
\end{tabular}}
\centering \caption{Performance comparison of algorithms on CIFAR100 and VireoFood-172}
\vspace{-0.25cm}
\label{tab:performance}
\end{center}
\end{table*}

\subsubsection{Datasets}
Experiments are conducted on two datasets: CIFAR-100LT and the more challenging VireoFood-172 \cite{chen2016deep} of 66,071 training and 33,154 test images.
\subsubsection{Evaluation Protocol}
For CIFAR-100LT dataset, we evaluated Top-1 accuracy under three different imbalance ratios: 100/50/10. For VireoFood-172, we evaluated Top-1 accuracy under an imbalance ratio of 50, where an imbalance ratio is defined as $N_{max}/N_{min}$. We followed TDE \cite{tang2020long} and CMLTNet \cite{li2024cross} to test the performance of the head, middle, and tail classes in the CIFAR100 dataset and VireoFood-172 dataset.
\subsubsection{Implementation Details}
For CIFAR100-LT,  we use warm-up scheduler for fair comparisons. All models were trained by using SGD optimizer with momentum µ = 0.9 and batch size 64. The learning rate was decayed by a cosine scheduler from 0.01 to 0.0 over 200 epochs for the ResNet50 and 40 epochs for the ViT and VPT . For  VireoFood-172-LT, all models were trained by using Adam optimizer with momentum µ = 0.1 and batch size 64. The learning rate is chosen in the range of 1e-4 to 5e-5. The learning rate decays every 4 epochs, with each model decaying 3 times by a factor of 0.1. 
\subsection{Performance Comparison}

We conducted a comprehensive comparison involving 3 visual modal backbones, 3 causal methods for image classification and 8 long-tailed methods: ResNet50 \cite{he2016deep}, ViT \cite{alexey2020image}, VPT\cite{jia2022visual}, CCIM \cite{yang2023context},  GOAT \cite{wang2024vision}, CaDeT \cite{pourkeshavarz2024cadet}, TDE \cite{tang2020long}, xERM \cite{zhu2022cross}, PLOT \cite{zhoupareto}, LiVT \cite{xu2023learning}, Gpaco \cite{cui2023generalized}, H2T \cite{li2024feature}, LPT \cite{dong2022lpt}. To make a fair comparison, the hyper-parameters of all models are chosen in above section. The following observations are drawn from Table \ref{tab:performance}:
\begin{itemize}
%TSCNet做到了模型无关，在多个不同的backbone上得到了性能提升，这源于TSCNet模型无关的反事实偏差校准策略，以及在patch和全局特征上对于相似语义偏差的去除提升了模型对于长尾数据的因果表示学习。

\item \textbf{TSCNet achieved significant improvements across different vision networks especially in the Transformers}. This is due to multi-level causal interventions that enhance the model's fine-grained causal representations, along with the introduction of a model-agnostic counterfactual bias calibration strategy.
%TSCNet 做到了比其他方法性能更高的提升。主要体现模型尾部类性能上的显著提升，在两个数据集上都达到了性能最高。
\item \textbf{TSCNet generally achieved better performance than other algorithms in both datasets.} The two-stage causal debiasing framework has been validated, demonstrating significant improvements in tail class performance while maintaining head class performance.

%用于图像分类的因果方法在头部类上的性能显著高于长尾算法。这是因为因果方法有效提升了数据丰富头部类的因果表示，对于尾部类的优化以及长尾分布的决策边界偏差去除仍然不足。
\item \textbf{Causal methods for image classification enhance head class performance but offer limited gains for tail classes}. They enhance causal representation for data-rich head classes, but fail to provide fine-grained causal representations for tail classes and do not address decision boundary bias in long-tail distributions.

%在长尾算法中，尾部类的提升显著，但是头部类的性能都存在下降。因果算法xERM仍然在ResNet上保持着优越的性能。
\item \textbf{Long-tail algorithms boost tail class performance but degrade head class.} Existing long-tail algorithms leads to a trade-off between the performance of head and tail classes due to spurious associations, and this is also the reason why they perform poorly on the imbalanced VireoFood-172 test set.

%长尾方法在Vireo172长尾数据集上总体性能提升并不如cifar显著，这主要源于Vireo172的测试集并不是平衡分布，这也显示了已有的长尾算法通过牺牲头部类性能来在平衡测试集上到达最佳性能的弊端。

\end{itemize}

\begin{table}[t]

\label{tab:module}
\small
\centering
\setlength{\tabcolsep}{0.8 mm}{\begin{tabular}{c|c|c|c|c|c|c}
\hline
\multirow{2}*{\bf {Models}} & \multicolumn{3}{c|}{\bf CIFAR100-ratio0.02} & \multicolumn{3}{c}{\bf VireoFood172-ratio0.02}\\
\cline{2-7}
{}  & {\bf Acc@all}& {\bf Acc@h}& {\bf Acc@t} & {\bf Acc@all} & {\bf Acc@h} & {\bf Acc@t}\\
\hline
ViT & 0.817  & 0.932 & 0.712  & 0.811  & 0.884 & 0.642 \\

+I & 0.839& \textbf{0.944} & 0.728 & 0.823 & 0.893 & 0.666 \\

+F & 0.828 & 0.942 & 0.713  & 0.820 & 0.892 & 0.665\\

+I+F & 0.845 & 0.943 & 0.748  & 0.826 & \textbf{0.893} &  0.667 \\

+I+F+C & 0.864  &0.933 & 0.805 & 0.839 & 0.878 & 0.741 \\

+I+F+C+R & \textbf{0.877}  & 0.937 & \textbf{0.819}  & \textbf{0.847} & 0.885  & \textbf{0.750}  \\

\hline
\end{tabular}}
\centering \caption{Ablation study of TSCNet with ViT backbone.}
\vspace{-0.25cm}
\label{tab:abliation}
\end{table}
\subsection{Ablation Study}
In this section, we further studied the working mechanism of each module of TSCNet, as shown in Table \ref{tab:abliation}. The following findings could be observed:

%加入patch-level可以显著提升ViT在长尾分布上的性能，尤其是对于尾部类上的性能的提升，同时feature-level的干预能进一步提升模型的泛化能力。
\begin{itemize}
\item{\verb||} \textbf{Hierarchical causal interventions(+I+F) effectively improve the feature representation for tail classes.} Incorporating patch-level intervention (+I) and feature-level intervention (+F) can significantly improve ViT's performance for tail classes, but there is still a significant gap compared to the head classes.

%加入反事实样本生成的平衡分布来训练模型，可以发现尾部类的性能上升明显，这源于对于数据分布的干预有效减少了长尾分布对于尾部类的干扰。

%加入反事实样本生成强度优化机制进一步改善了模型对于尾部类的学习，同时保持了对于头部类的性能
\item{\verb||} \textbf{Counterfactual generation (+C) helps further optimize the decision boundary.} Training the model with a counterfactual balanced distribution (+C) leads to a significant improvement in tail class performance. However, the uncontrollable counterfactual strength leads to performance decline in the head classes.
% With the assistance of the pre-learned hierarchical knowledge base (+K), the filtered tags achieve better classification performance. It is comparable to the basic model in Ingredient-101 which verifies the effectiveness of introducing VSK to deal with noise. Although the improvement of (+L+K) is not obvious on the NUS-WIDE dataset which contains complex semantic information and has heavier noise in cross-modal inference, the boosting in performance after combining with the heterogeneous graph (+G) is still comparable. 
\item{\verb||} \textbf{Counterfactual refinement (+R) enhances learning for both head and tail classes.} By adopting a difficulty-progressive refining strategy (+R), it simulates different data distributions to finely model the spurious correlations between head and tail logits.
\end{itemize}

\subsection{In-depth Analyses}
\begin{table}[t]

\label{tab:module}
\small
\centering
\setlength{\tabcolsep}{0.8 mm}{\begin{tabular}{c|c|c|c|c|c|c}
\hline
\multirow{2}*{\bf {Setting}} & \multicolumn{3}{c|}{\bf CIFAR100-ratio0.02} & \multicolumn{3}{c}{\bf CIFAR100-ratio0.02}\\
\cline{2-7}
{}  & {\bf Acc@all}& {\bf Acc@h}& {\bf Acc@t} & {\bf Acc@all} & {\bf Acc@h} & {\bf Acc@t}\\
\hline
Base & 0.817 & 0.932 & 0.712  & 0.811  & 0.884 &  0.642 \\

Random & 0.821 &0.938 & 0.702  & 0.806 & 0.881 & 0.642 \\

Zero & 0.816 & 0.937 &  0.688 & 0.810 & 0.881  & 0.642 \\

Average &0.820  & 0.938 & 0.703 & 0.810 & 0.887 & 0.637 \\

Confounder &\textbf{0.828} &\textbf{0.942}&\textbf{0.713} & \textbf{0.826}  & \textbf{0.896} & \textbf{0.657}  \\
\cline{1-7}

\hline
\end{tabular}}
\centering \caption{The results on different versions of dictionary in Feature-level Invention model. Random, Zero, Average Feature, and Confounder Dictionary use random features, zero features, the average features of the image, and the average features of the class-independent factors as confounders.}
\vspace{-0.25cm}
\label{tab:aaaa}
\end{table}

\subsubsection{Effectiveness Analyses of the Dictionary $S_p$}
As delineated in Table \ref{tab:aaaa}, we scrutinized the efficacy of the class-non confounder dictionary $S_p$ within the BD module. Experimental results show that replacing $S_p$ with a random dictionary or a zero dictionary significantly deteriorates performance. Using average Image features as a confounder dictionary is less effective than class-agnostic confounder features, indicating that random dictionaries and class averages, among others, are insufficient as confounder.
\begin{figure}[t]
\centering
\includegraphics[width=0.49\textwidth]{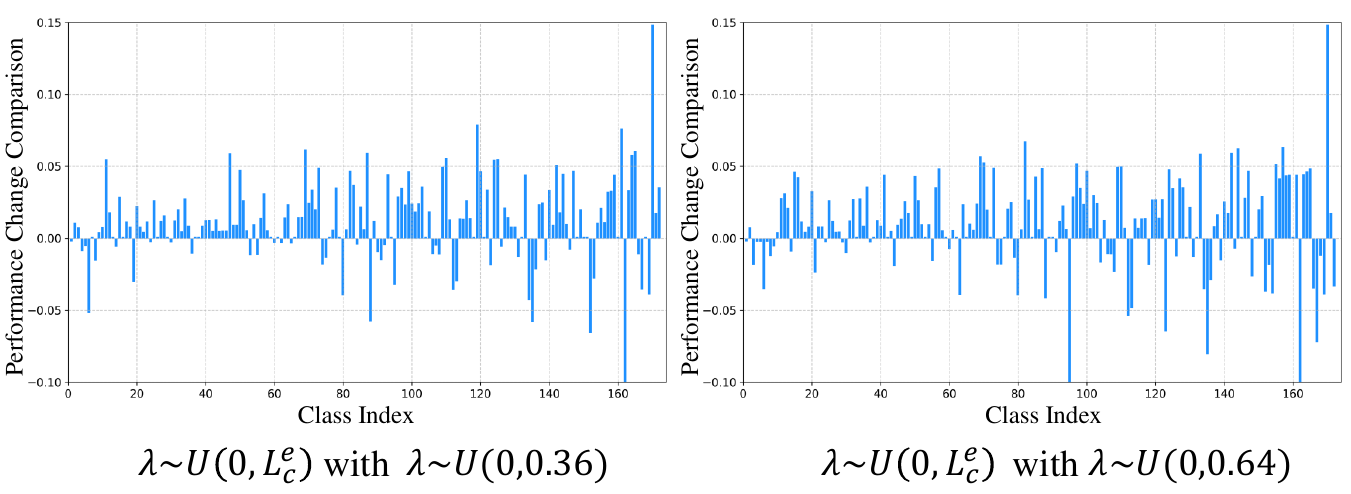}
\caption{Comparison of our adaptive adjustment of the enhancement parameter $L_c^e$  with using fixed parameters 0.36 and 0.64 in terms of performance, positive values indicate that our parameter adjustment performs better in this category.}
\label{fig:fig5}
\vspace{-0.25cm}
\end{figure}
\subsubsection{Effect of the counterfactual enhancement parameter $L_c^e$}
As shown in Figure \ref{fig:fig5}, compared to the multiple fixed parameters (0.36, 0.64), the adaptive adjustment of the enhancement parameter $L_c^e$ achieves performance improvements on most categories of the VireoFood-172 dataset. This effectively demonstrates that the designed adaptive enhancement parameter $L_c^e$ can progressively increase the counterfactual data strength from easy to difficult and assign appropriate enhancement intensity to each class. 
\subsection{Case Study}

\subsubsection{Visualization of the Causal Representation by HCRL
}
\begin{figure}[t]
\centering
\includegraphics[width=0.49\textwidth]{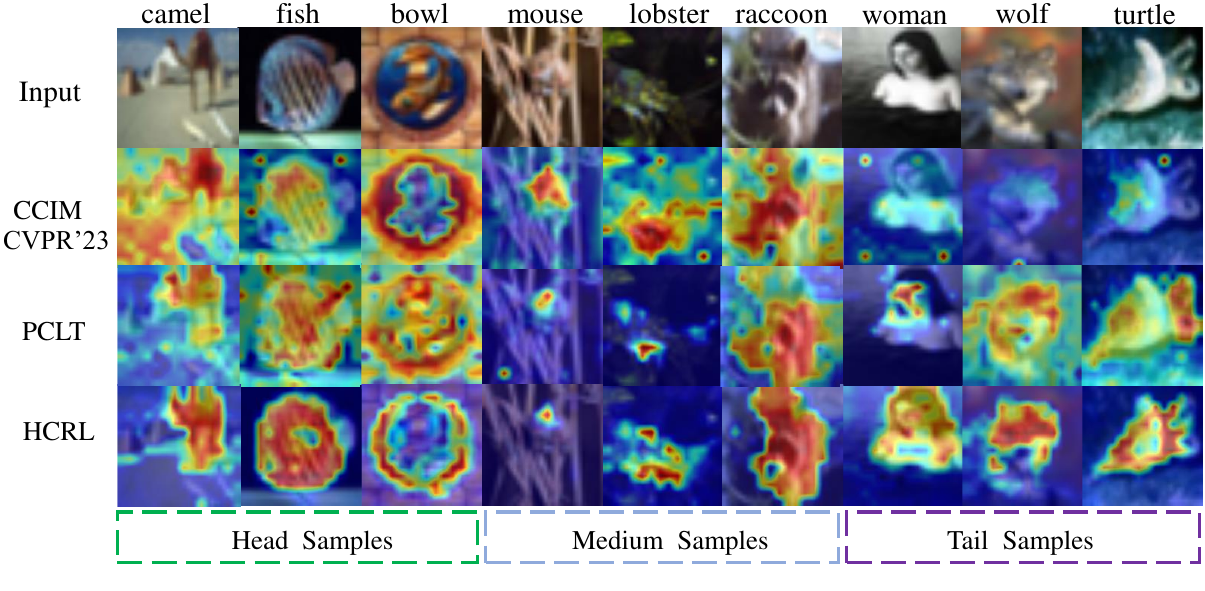}
\caption{Visualization of attention on sampled image from the CIFAR100-LT. Four lines respectively represent the input, the CCIM, the Patch-level Intervention (PCLT) and HCRL.}
\label{fig:fig6}
\vspace{-0.25cm}
\end{figure}

To assess the efficacy of HCRL, we conducted a comparative evaluation against CCIM \cite{liu2022contextual} on the CIFAR100 validation set, emphasizing causal visual information in long-tail data using GradCAM \cite{selvaraju2020grad} heatmaps as shown in Figure \ref{fig:fig6}. CCIM demonstrated weak causal perception for tail-class images such as "wolf" and "woman." In contrast, the proposed patch-level intervention method (PCLT) enriched fine-grained causal representations, enabling the model to capture intricate features, such as facial details in "woman" and "mouse," as well as edge features in "flowerpot," effectively mitigating the interference of irrelevant information. Furthermore, HCRL improved the model’s attention, enhancing its intervention on tail-class.

\subsubsection{Visualization of the Decision Boundary by CLBC}
\begin{figure}[t]
\centering
\includegraphics[width=0.49\textwidth]{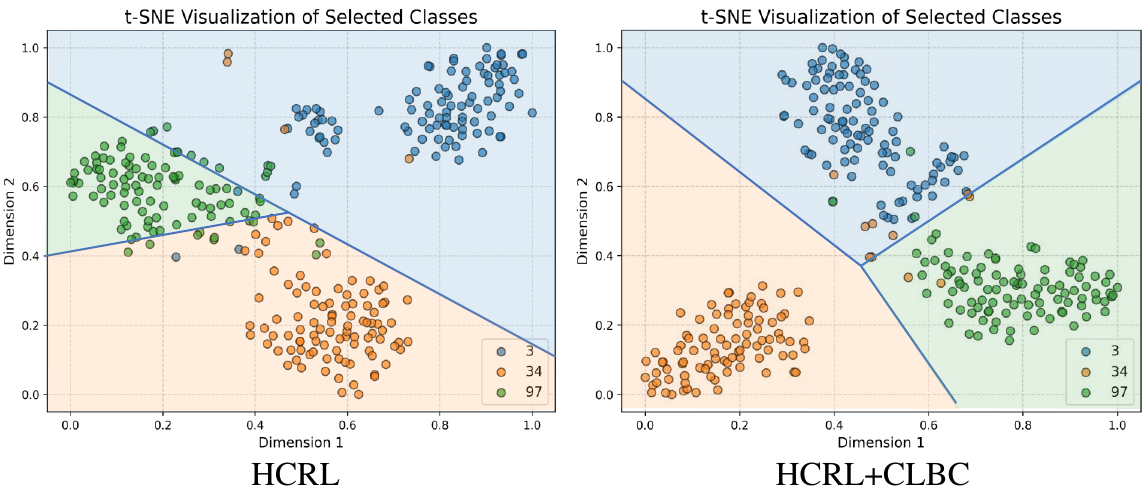}
\caption{ Visualization results of HCRL and HCRL+CLBC on
the CIFAR100 dataset. The regions in different colors represent the predictions for the corresponding color categories.}
\label{fig:fig7}
\vspace{-0.25cm}
\end{figure}
We utilized tSNE visualization to illustrate the feature distribution and decision boundaries of three commonly confused head-tail categories in the CIFAR100 dataset, as shown in Figure \ref{fig:fig7}. We observed a clearly separable boundary for the three categories in the feature space through HCRL. However, due to the confounding effect of the data distribution, a significant number of tail-class samples were misclassified as head-class samples. By incorporating the CLBC module and constructing a counterfactual balanced distribution, the model's bias towards tail classes at the decision boundary was effectively mitigated. Notably, the CLBC module further enhanced the model's representation of long-tail data.

\subsubsection{Error confusion analysis of TSCNet}
\begin{figure}[t]
\centering
\includegraphics[width=0.49\textwidth]{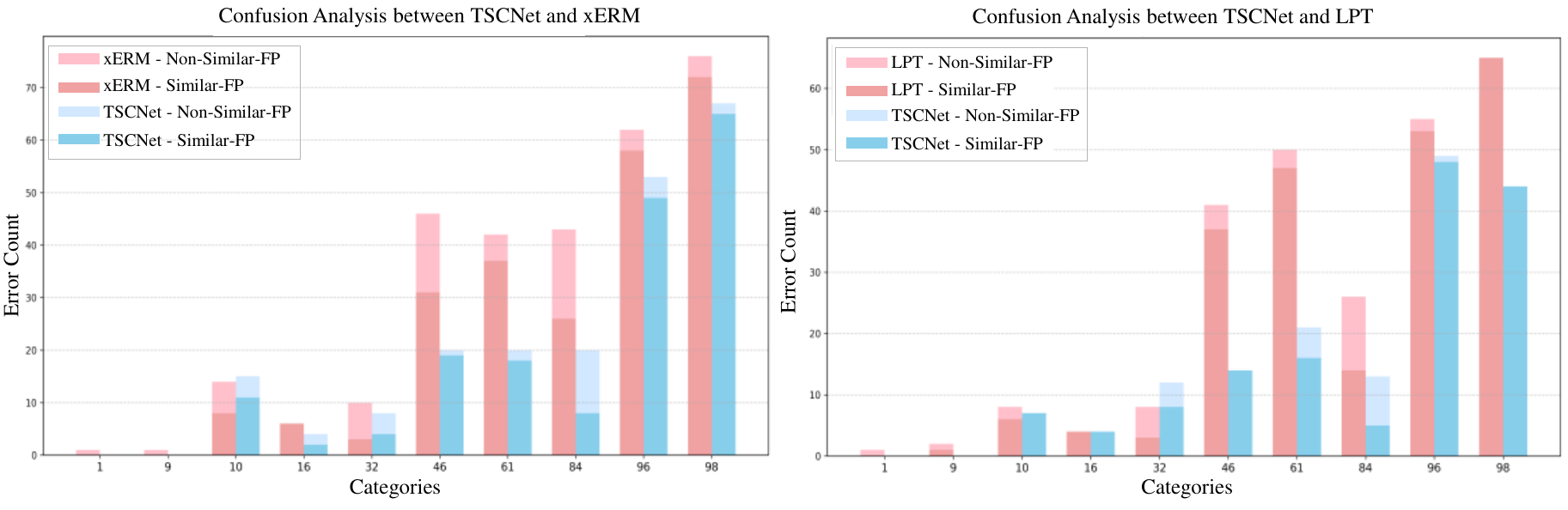}
\caption{The error confusion analysis of TSCNet, and the comparison of xERM and LPT in terms of similar-FP and non-similar-FP in CIFAR100.}
\label{fig:fig8}
\vspace{-0.25cm}
\end{figure}
We compared the error confusions of different models with our method, as shown in Figure \ref{fig:fig8}. The error confusions were categorized into two types: similar-class confusion and non-similar-class confusion. The results demonstrate that TSCNet effectively alleviates error confusions in similar classes by leveraging fine-grained causal representations and bias calibration, particularly in the middle and tail classes, with a more significant improvement observed in the middle classes. In contrast, xERM and LPT faces challenges in modeling the relationship between feature regions in ViT and model predictions, leading to a concentration of errors in similar classes.

\subsection{Conclusion}
To effectively mitigate the biases induced by long-tail distributions and tackle the challenges associated with applying existing causal methods to ViT, this paper introduces TSCNet. Specifically, TSCNet strengthens the model's fine-grained causal representation through hierarchical causal representation learning. Furthermore, TSCNet employs a model-agnostic counterfactual logits bias calibration stage to adaptively eliminate the prediction biases caused by long-tail distributions. Experimental results indicate that the synergistic interaction of the two stages significantly enhances long-tailed image classification performance across various backbones. Therefore, future work will focus on exploring the use of causal methods in large models to further improve long-tail image classification performance.

\bibliographystyle{named}
\bibliography{ijcai25}

\end{document}